\documentclass{article}

\usepackage{iclr2027_conference,times}

\usepackage{amsmath,amsfonts,bm}

\def\eqref#1{equation~\ref{#1}}

\def\1{\bm{1}}

\DeclareMathAlphabet{\mathsfit}{\encodingdefault}{\sfdefault}{m}{sl}
\SetMathAlphabet{\mathsfit}{bold}{\encodingdefault}{\sfdefault}{bx}{n}

\usepackage{hyperref}
\usepackage{url}
\usepackage{graphicx}
\usepackage{capt-of}
\usepackage{booktabs}
\usepackage{array}
\usepackage{multirow}
\usepackage{amsmath,amssymb,mathtools}
\newtheorem{tbproposition}{Proposition}[section]
\newenvironment{tbproof}{\par\noindent\textit{Proof.}\ }
  {\hfill\(\square\)\par}
\usepackage{xcolor}
\usepackage{colortbl}
\usepackage{microtype}
\usepackage{fvextra}
\usepackage{placeins}
\usepackage{enumitem}
\usepackage{float}
\usepackage{algpseudocode}
\floatstyle{ruled}
\newfloat{algorithm}{tbp}{loa}
\floatname{algorithm}{Algorithm}
\graphicspath{{./}}

\newcommand{\method}{\textsc{TeachBack}}

\newcommand{\gain}[1]{\textcolor{green!50!black}{\scriptsize(#1)}}
\DefineVerbatimEnvironment{promptverbatim}{Verbatim}{%
  fontsize=\scriptsize,
  breaklines=true,
  breakanywhere=true,
  breakautoindent=false,
  breakindent=0pt,
  breaksymbolleft={},
  breaksymbolright={}}

\title{\method: Learning to Teach via Bidirectional Preference Optimization\\
for On-Policy Self-Distillation}
\author{%
\parbox{\dimexpr\textwidth-2\tabcolsep\relax}{%
\centering\normalfont
\fontsize{11}{15}\selectfont
Linfei Ji$^{1}$\quad Jiayun Li$^{2}$\quad
Haozhao Wang$^{1}$\quad Siyu Jiang$^{1}$\\[1pt]
Zehao Wang$^{3}$\quad Longtao Xiao$^{1}$\quad
Yu-Jie Yuan$^{4}$\thanks{\raggedright Corresponding authors: Yu-Jie Yuan (\href{mailto:yuanyujie66@gmail.com}{\texttt{yuanyujie66@gmail.com}});\\
Ruixuan Li (\href{mailto:rxli@hust.edu.cn}{\texttt{rxli@hust.edu.cn}}).}\quad
Ruixuan Li$^{1}$\footnotemark[1]\\[7pt]
\fontsize{9}{12}\selectfont
$^{1}$\,Huazhong University of Science and Technology\\
$^{2}$\,Tsinghua University\qquad
$^{3}$\,Harbin Institute of Technology\\
$^{4}$\,Huawei Technologies Ltd.\endgraf
}%
}
\iclrfinalcopy

\begin{document}
\maketitle
\lhead{}
\renewcommand{\headrulewidth}{0pt}

\begin{abstract}
On-policy self-distillation (OPSD) enables a language model to learn from a Teacher instantiated from the same model but given access to privileged information, such as a reference solution. We find that access to sufficient privileged information allows the Teacher to generate highly accurate trajectories; however, accuracy alone does not guarantee that these trajectories are suitable for Student learning. Direct access to a complete solution may encourage shortened, reference-conditioned derivations that omit the exploration, verification, and self-correction required to solve problems without privileged information.
We introduce TeachBack, a bidirectional preference-learning framework for on-policy self-distillation that learns not only from the Teacher but also how the Teacher should teach. TeachBack shapes the Teacher’s generation behavior by controlling the length of the reference-reasoning suffix injected into the prompt and by varying the prompt style. While maintaining comparable Teacher accuracy, it encourages the Teacher to explore and re-solve problems, thereby reducing its reliance on the reference solution. At each rollout refresh, the shared policy generates both a privileged Teacher trajectory and a problem-only Student trajectory. A pairwise preference objective uses privileged Teacher trajectories to improve Student learning, while the reversed preference uses the Student’s behavior to shape future Teacher supervision under the privileged condition. Experiments on mathematical and logical reasoning tasks show that TeachBack consistently improves Student performance over existing OPSD methods and related baselines.
\end{abstract}

\section{Introduction}
\label{sec:introduction}
On-policy distillation~\citep{agarwal2024policy}, which provides dense supervision by allowing a Student to sample reasoning trajectories from its current policy while a stronger Teacher evaluates the Student-generated prefixes, has emerged as the dominant post-training paradigm for enhancing the reasoning capabilities of large language models. Building on this, On-policy self-distillation (OPSD)~\citep{zhao2026self} further removes the need for a separate Teacher by employing a unified, parameter-shared policy, receiving great attention recently. This policy allows the model to play two roles, where the model acts as a Student when given only the question, but switches to a privileged-information (PI) Teacher when provided with additional reasoning evidence, such as reference solutions. Despite eliminating the dependency on external models, recent studies reveal that OPSD may cause shorter responses and a degeneration of native reasoning behaviors, a failure mode commonly referred to as \textit{thinking collapse}~\citep{peng2026diagnosing}, which is particularly pronounced in models relying on long chain-of-thought reasoning.

Several approaches have been proposed to address this degradation. \textit{One representative line} studies OPSD from the perspective of optimization, where an uncertain Student and a PI-conditioned Teacher can favor different continuations at high-entropy reasoning forks. More specifically, privileged targets can suppress the Student's branching, verification, and backtracking when their token probabilities are directly matched~\citep{kaur2026rethinking}. To preserve these behaviors, AD-OPSD~\citep{peng2026diagnosing} anchors updates at high-risk tokens to a frozen base policy. \textit{Another line} examines the supervision signal itself: reference conditioning can change the Teacher's response style without providing correspondingly useful reasoning. RLCSD~\citep{pan2026rlcsd} contrasts correct and incorrect hints to cancel style shifts unrelated to correctness, while Purified OPSD~\citep{shen2026purified} removes a reference-only component from the Teacher signal. \textit{The third line} changes how the Student learns from that signal; PBSD~\citep{yu2026preference} uses reward-regularized preferences between Teacher and Student rollouts instead of directly matching their token distributions. Although these methods have made great achievements, \textit{they primarily regulate how the Student uses a Teacher signal that has already been generated}, leaving open a more upstream question: \textit{can PI be used to construct a Teacher whose reasoning remains useful to a Student that cannot access PI?}
\begin{figure}[H]
\centering
\includegraphics[width=\linewidth]{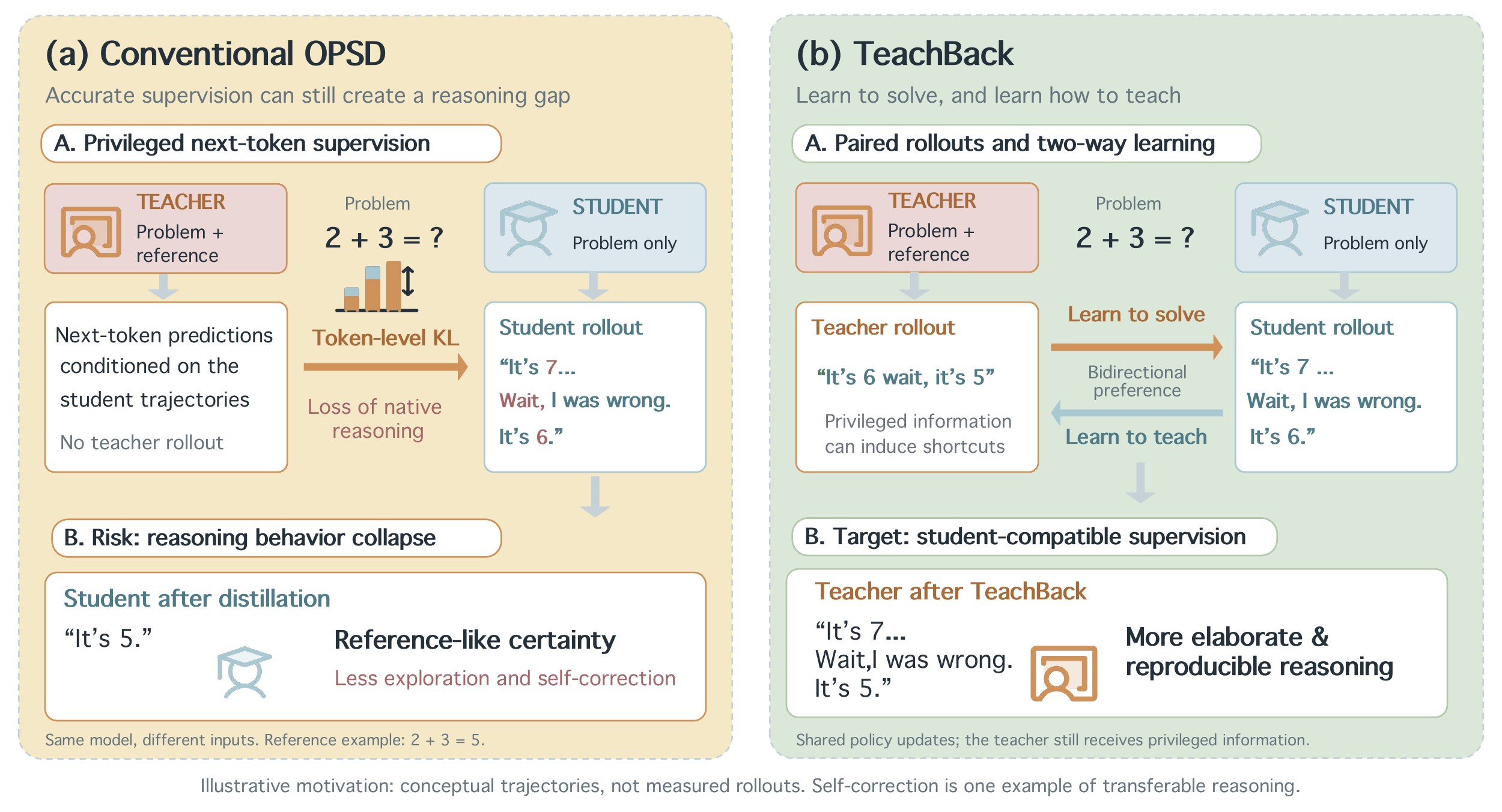}
\caption{Motivation for \method. In standard OPSD, access to privileged information can lead the Teacher to produce a short solution that omits exploration and verification. \method{} uses partial reference reasoning and Student feedback to encourage complete, self-contained Teacher trajectories that remain useful to a problem-only Student.}
\label{fig:motivation}
\end{figure}

This question points to a Teacher-side way to mitigate thinking collapse: privileged information should guide the Teacher's reasoning rather than serve as an authoritative solution to be copied. A problem-only Student must search among alternative paths, express uncertainty at ambiguous steps, verify intermediate conclusions, and recover from mistakes. As we design in Figure~\ref{fig:motivation}, a useful PI-conditioned Teacher should preserve these Student-compatible behaviors while using privileged information to check the explored path, correct unproductive branches, and converge more reliably to a correct answer. The desired Teacher therefore needs to balance three properties: high correctness under privileged information, low dependence on the wording or trajectory of the reference solution, and productive exploration that remains reproducible without PI. In particular, the Teacher should not simply compress the reference-conditioned response into a shorter answer. It should re-solve the original problem, use privileged information for verification, and produce a complete trajectory that is both reliable as supervision and rich enough for the Student to learn.  

Based on this motivation, we propose \method, which improves both the construction and transfer of privileged Teacher trajectories. First, \method{} presents the Teacher with partial reference reasoning of varying lengths and prompt styles as optional working notes, then asks it to solve the original problem from the beginning. This encourages the Teacher to use PI to check its reasoning without simply following the reference. The shared policy then generates a Teacher trajectory with PI and a Student trajectory from the problem alone. We train with preferences in both directions: the Teacher trajectory guides Student learning, while the Student trajectory provides feedback on which reasoning paths the Teacher should retain. In this way, the Teacher learns to produce accurate, less reference-dependent supervision that the Student can learn from.

Our contributions are summarized as follows:

\begin{itemize}[leftmargin=*,nosep]

\item Through explicit Teacher rollouts, we reveal a substantial
reasoning-behavior gap between the PI-conditioned Teacher and the
problem-only Student, and show that highly accurate Teacher trajectories
are not necessarily suitable for Student learning.

\item We construct the PI-conditioned Teacher by presenting a controllable
suffix of reference reasoning as optional working notes and varying the
surrounding prompt style. This encourages the Teacher to re-solve the problem
and produce complete reasoning with less reliance on the reference.

\item We develop a bidirectional preference-learning framework over paired
Teacher and Student trajectories. Teacher-origin trajectories guide Student
learning, while Student-origin trajectories provide compatibility feedback
that shapes future Teacher supervision.

\item Experiments on mathematical and logical reasoning tasks show that
\method{} consistently improves Student's reasoning performance and
outperforms existing OPSD and related baselines.

\end{itemize}

\section{Related Work}
\label{sec:related-work}
\paragraph{On-policy distillation and privileged supervision.}
Classical knowledge distillation~\citep{hinton2015distilling,kim2016sequence,gu2024minillm} transfers a
stronger Teacher's predictions or generated sequences to a Student.
On-policy distillation (OPD)~\citep{agarwal2024policy} instead queries the
Teacher on Student-sampled prefixes, reducing the mismatch between training
and deployment states. On-policy self-distillation
(OPSD)~\citep{zhao2026self} uses one parameter-shared policy as a problem-only
Student and a privileged-information (PI) Teacher, and Privileged Information
Distillation~\citep{penaloza2026privileged} extends this setting to long-horizon
agents~\citep{wang2026skill,wu2026seed,lu2026self}. Recent studies show that privileged conditioning can suppress epistemic
verbalization, exploration, verification, and backtracking at high-entropy
reasoning forks, leading to thinking collapse and reference-induced style
drift~\citep{kim2026does,kaur2026rethinking,peng2026diagnosing,pan2026rlcsd}.
Existing remedies modify the resulting Teacher signal: Purified
OPSD~\citep{shen2026purified} removes the reference-induced component;
PAINT~\citep{tan2026paint}, Adaptive Teacher
Exposure~\citep{han2026adaptive}, and best-of-$N$
selection~\citep{zhang2026policy} control privileged exposure or retain
selected rollouts; and AR-OPD~\citep{zhang2026beyond} and
DOPD~\citep{yu2026dopd} constrain or route privileged supervision according
to local compatibility and relative advantage. These methods improve how an
already formed signal is filtered or transferred, leaving the construction of
a Student-compatible Teacher as an upstream problem.

\paragraph{Preference optimization and Teacher construction.}
Preference optimization provides a sequence-level alternative to direct
token-distribution matching. DPO~\citep{rafailov2023direct} learns a
reference-relative policy from response preferences, while
SPIN~\citep{chen2024self} uses self-generated responses to form an iterative
preference signal. PBSD~\citep{yu2026preference} applies preference learning
to on-policy self-distillation by optimizing a reward-regularized preference
gap between PI-conditioned Teacher samples and on-policy Student samples
instead of directly matching Teacher probabilities. Related methods use
consensus gating~\citep{stein2026gates} or
self-revision~\citep{he2026self,zelikman2022star} to select or construct supervision~\citep{li2026policy} before
distillation. TeachBack addresses the remaining Teacher-construction problem
by presenting a controllable ground-truth suffix as incomplete working notes,
varying the prompt framing, and using two cross-condition preference
directions: the Student-side objective transfers useful Teacher trajectories
under the problem-only condition, while the reversed objective uses Student
behavior to shape future privileged Teacher trajectories.

\section{Method}
\label{sec:method}
\begin{figure}[t]
\centering
\includegraphics[width=\linewidth]{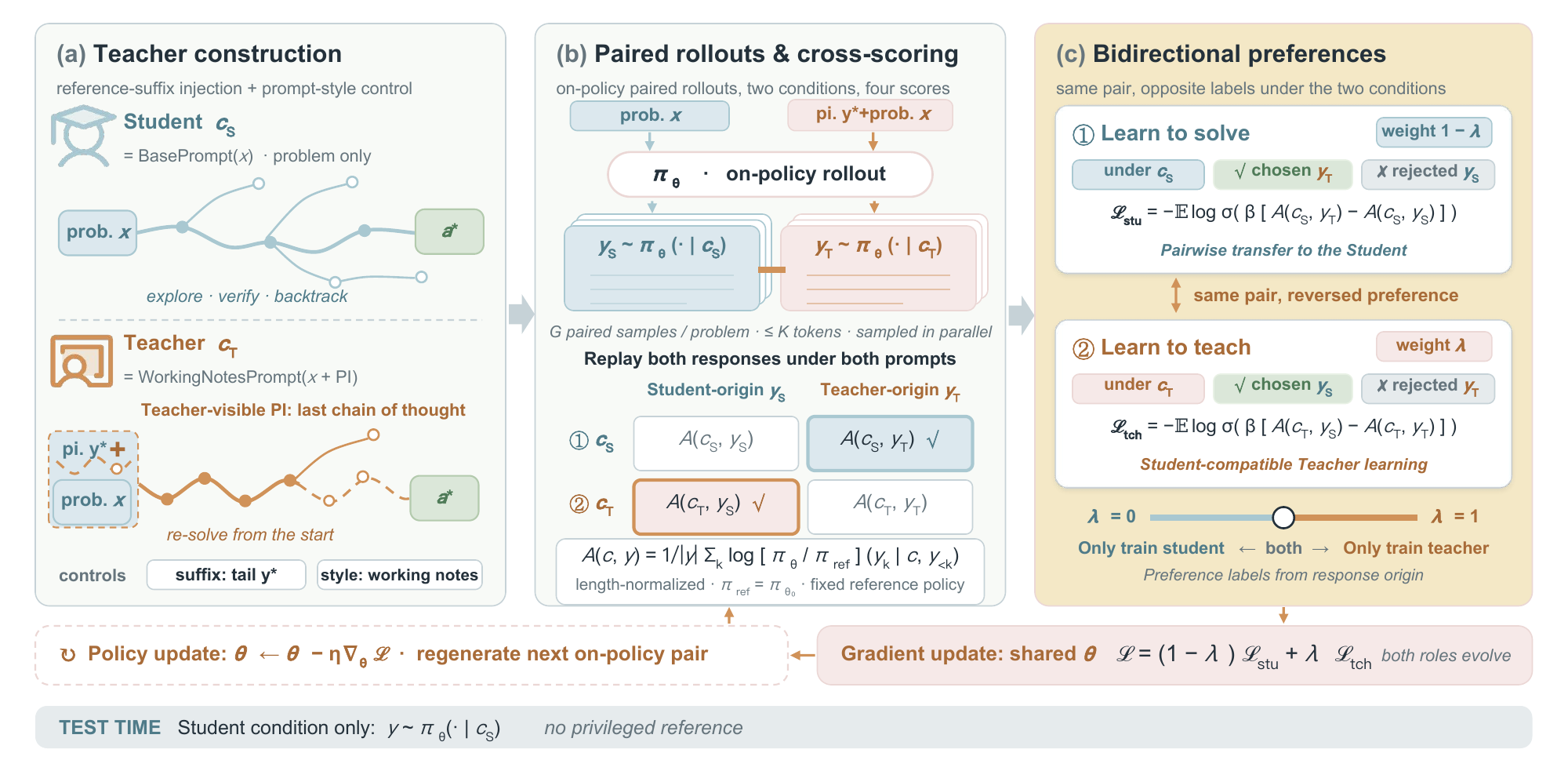}
\caption{Overview of \method{}. (a) The Student receives the problem alone, while a reference-reasoning suffix is framed as optional working notes for the Teacher to re-solve the problem. (b) The shared policy samples paired Student and Teacher trajectories; both responses are replayed under both conditions and scored against a fixed reference policy. (c) Under the problem-only condition, the Teacher-origin response is preferred; under the PI-conditioned condition, the Student-origin response is preferred. The weighted losses update the shared policy before the next on-policy pair is generated. At test time, only the problem-only Student condition is used.}
\label{fig:overview}
\end{figure}

Given training examples \(\mathcal D=\{(x_i,y_i^*,a_i^*)\}_{i=1}^{N}\), where \(x_i\) is a problem, \(y_i^*\) is its reference reasoning, and \(a_i^*\) is its correct answer, TeachBack uses a suffix of \(y_i^*\) as privileged information (PI) and \(a_i^*\) for answer verification. The Student and Teacher are conditional roles of one parameter-shared policy \(\pi_\theta\), rather than separate networks: the Student receives only the problem, while the Teacher receives the problem and a prompt containing PI. Training alternates paired rollout generation with bidirectional preference updates. Figure~\ref{fig:overview} summarizes Teacher construction, paired rollouts, cross-scoring, and the two preference updates.

\subsection*{3.1 Teacher Construction: Reference-Suffix Injection and Prompt-Style Control}
\addcontentsline{toc}{subsection}{3.1 Teacher Construction: Reference-Suffix Injection and Prompt-Style Control}

For each problem \(x\), the Student context and PI-conditioned Teacher context are
\begin{equation}
 c_S=\operatorname{BasePrompt}(x),\qquad
 c_T=\operatorname{WorkingNotesPrompt}\!\left(x,\operatorname{Tail}(y^*)\right).
\label{eq:teacher-construction}
\end{equation}
Equation~\ref{eq:teacher-construction} shows the Working Notes configuration used in the main training runs. The suffix is extracted from the dataset reference rather than from the current Student response, so Student sampling errors do not change the privileged information in the Teacher prompt. We use a suffix instead of the complete reference because it provides answer-relevant information while leaving the Teacher responsible for reconstructing the earlier derivation. An authoritative solution framing may encourage the Teacher to continue or copy the suffix. Framing it as incomplete, non-terminal working notes instead prompts the Teacher to re-solve the problem, verify the notes, fill in missing steps, and produce a complete answer.

We study two related factors in Teacher construction. The suffix length controls how much privileged information is available, while the prompt style changes how that information is presented. We compare several styles, including authoritative reference, answer-only hint, and optional working notes; the answer-only and full-solution instruction styles also change the information content. These comparisons let us assess Teacher correctness and reasoning behavior separately from the downstream preference objective. Appendix~\ref{app:working-notes-prompt} gives the audit setup, metric definitions, and prompt templates.

\subsection*{3.2 Paired On-Policy Rollouts and Cross-Scoring}
\addcontentsline{toc}{subsection}{3.2 Paired On-Policy Rollouts and Cross-Scoring}

Unlike conventional OPSD, which transfers privileged next-token predictions on Student-generated prefixes, TeachBack generates explicit trajectories from both conditional roles. At refresh \(t\), the current policy is held fixed while generating \(G\) paired responses:
\begin{equation}
 y_{S,i}^{(t,g)}\sim\pi_{\theta_t}(\cdot\mid c_{S,i}),\qquad
 y_{T,i}^{(t,g)}\sim\pi_{\theta_t}(\cdot\mid c_{T,i}),
 \quad g=1,\ldots,G.
\label{eq:paired-rollouts}
\end{equation}
Both responses are sampled from the same policy state, so systematic differences between their rollout distributions arise from the prompt conditions rather than separate models. We retain the first \(K\) valid response tokens for preference training; below, \(y_S\) and \(y_T\) denote these retained prefixes when used in scoring equations, with example and rollout indices omitted for readability. This prefix focuses the objective on the part of the trajectory where the model establishes a solution direction, explores alternatives, and begins verification; later tokens often elaborate a path that has already been selected. Generation budgets, truncation rules, and the longer diagnostic rollout cap are specified in Appendix~\ref{app:hyperparameters}.

The sampled responses are replayed under both prompts:
\begin{equation}
\mathcal Q^{(t)}=
\begin{bmatrix}
 c_S\oplus y_S^{(t)} & c_S\oplus y_T^{(t)}\\
 c_T\oplus y_S^{(t)} & c_T\oplus y_T^{(t)}
\end{bmatrix}.
\label{eq:cross-scoring}
\end{equation}
The current policy and a frozen initialization policy score the same response tokens in all four combinations. Within each row, the scoring condition is fixed and the response origin varies. In particular, \(y_T\) replayed under \(c_S\) lets us evaluate Teacher reasoning under the problem-only condition. Replaying \(y_S\) under \(c_T\) supplies the corresponding compatibility view for the Teacher role. After the preference update, the new policy generates the next paired batch.

\subsection*{3.3 Bidirectional Preference Optimization}
\addcontentsline{toc}{subsection}{3.3 Bidirectional Preference Optimization}

Let \(\pi_{\mathrm{ref}}=\pi_{\theta_0}\) remain fixed throughout training. For condition \(c\) and response \(y\), we use the length-normalized score relative to this frozen reference policy:
\begin{equation}
 A_\theta^c(y)=\frac{1}{|y|}\sum_{k=1}^{|y|}
 \log\frac{\pi_\theta(y_k\mid c,y_{<k})}
 {\pi_{\mathrm{ref}}(y_k\mid c,y_{<k})}.
\label{eq:response-score}
\end{equation}
Only valid response positions contribute. Length normalization prevents longer responses from receiving a larger score solely because they contain more log-probability terms, while the frozen reference anchors both conditional comparisons to the same initialization.

\paragraph{Student-side transfer.}
Under the problem-only condition, we prefer the Teacher-origin response, \(y_T\succ y_S\), using
\begin{equation}
\mathcal L_{\mathrm{stu}}
=-\mathbb E\log\sigma\!\left(
\beta\left[A_\theta^{c_S}(y_T)-A_\theta^{c_S}(y_S)\right]\right).
\label{eq:student-pair}
\end{equation}

Here, $\sigma$ is the logistic sigmoid, $\beta>0$ scales the preference
margin, and the expectation is over training problems and paired rollouts
sampled from the fixed policy snapshot at the current refresh. We use
$\beta=0.1$.
Both responses are scored under \(c_S\). The margin therefore asks whether the trajectory generated with privileged information becomes preferable under the deployment condition. This transfers a trajectory-level teaching signal without requiring the Student to match the complete PI-conditioned next-token distribution. The label is determined by response origin, not by assuming that every Teacher sample is correct; Teacher correctness is measured separately in the experiments.

\paragraph{Teacher-side feedback.}
Under the PI-conditioned condition, we reverse the ordering and prefer the Student-origin response, \(y_S\succ y_T\):
\begin{equation}
\mathcal L_{\mathrm{tch}}
=-\mathbb E\log\sigma\!\left(
\beta\left[A_\theta^{c_T}(y_S)-A_\theta^{c_T}(y_T)\right]\right).
\label{eq:teacher-pair}
\end{equation}

The reversed preference does not claim that the Student answer is more accurate. A problem-only response can contain incremental derivation, verification, and backtracking that are absent from a concise privileged response. Favoring Student-origin responses under \(c_T\) can feed Student-compatible reasoning back into the role that generates future Teacher targets, while the privileged suffix continues to support correctness. Because both roles share parameters, the Teacher-side update changes the next privileged rollout as well as the Student. It can also reinforce Student errors or unproductive verbosity, which is why we evaluate correctness, exploration, reference dependence, response length, and downstream performance together. The numerical margins and their training dynamics are analyzed in Section~4.4, and the local compatibility analysis is given in Appendix~\ref{app:theoretical-analysis}.

\subsection*{3.4 Joint Co-adaptation and Training Protocol}
\addcontentsline{toc}{subsection}{3.4 Joint Co-adaptation and Training Protocol}

The two directions are combined as
\begin{equation}
\mathcal L_{\mathrm{main}}^{(t)}(\lambda)
=(1-\lambda)\mathcal L_{\mathrm{stu}}^{(t)}
+\lambda\mathcal L_{\mathrm{tch}}^{(t)},
\qquad \lambda\in[0,1].
\label{eq:new-joint}
\end{equation}
The boundary \(\lambda=0\) isolates Teacher-to-Student transfer, \(\lambda=1\) isolates Teacher-side feedback, and intermediate values test co-adaptation. Each refresh generates paired trajectories from the current policy, replays both responses under both conditions, computes the two preference losses, and updates only the shared parameters. Rollouts are stop-gradient and \(\pi_{\mathrm{ref}}\) remains frozen; verifier outcomes do not filter pairs or rescale the losses. Algorithm~1 in Appendix~\ref{app:algorithm} gives the complete procedure, while Appendix~\ref{app:hyperparameters} lists the fixed budgets, masks, and refresh settings.

\section{Experimental Protocol}
\label{sec:experiments}
We organize the experiments as a progression of related questions. We first
ask whether TeachBack improves mathematical and logical reasoning when only the
original problem is available at deployment. We then examine which Teacher
prompt style and privileged-information ratio best balance correctness,
reasoning elaboration, response length, and dependence on the reference. Next,
we test whether bidirectional preference optimization moves the Student and
Teacher margins in the intended directions while preserving stable entropy and
gradient behavior. Finally, we study how the Teacher-side objective weight
affects training dynamics and final performance.

\subsection{Experimental Setup}

\textbf{Models and data.} We train Qwen3-1.7B, Qwen3-4B, and Qwen3-8B~\citep{yang2025qwen3} on a challenging subset of DeepMath~\citep{deepmath}. We evaluate on AMC 2023~\citep{maa2023amc} and AIME 2024/2025~\citep{aime24,aime25}, and on Knight-and-Knaves (KK)~\citep{xie2025memorization}, using 4--8 Roles in-domain and 9--11 Roles out-of-domain. Cross-architecture Math results on DeepSeek-R1-Distill-Qwen-7B and DeepSeek-R1-Distill-Llama-8B~\citep{guo2025deepseek} are reported in Appendix~\ref{app:cross-model-results}. The dataset splits, training settings, and evaluation protocol are detailed in Appendix~\ref{app:hyperparameters}.

\textbf{Baselines.}
We compare TeachBack with Base, GRPO~\citep{shao2024deepseekmath}, OPSD~\citep{zhao2026self}, SPIN~\citep{chen2024self}, and PBSD~\citep{yu2026preference}. SPIN uses the dataset ground-truth reasoning trace as the preferred response against model-generated responses. PBSD uses trajectories generated by a frozen privileged teacher as preferred responses against on-policy student trajectories. TeachBack generates both trajectories from the current shared policy and optimizes preferences under both prompt conditions. 
\subsection{Main Results}

\begin{table*}[t]
\caption{Main results across model scales. Mathematical reasoning is reported as mean@12. Knight-and-Knaves (KK) is reported as pass@1, with 4--8 Roles as the in-domain (ID) setting and 9--11 Roles as out-of-domain (OOD) settings. Bold and underlined entries denote the best and second-best results within each model block; parenthetical values are TeachBack gains over Base.}
\label{tab:main-results-current}
\begin{center}
\setlength{\tabcolsep}{3.2pt}
\renewcommand{\arraystretch}{1.03}
\scriptsize
\resizebox{\textwidth}{!}{%
\begin{tabular}{llccccccccc}
\toprule
& & \multicolumn{4}{c}{\textbf{Mathematical Reasoning} (mean@12)}
& \multicolumn{5}{c}{\textbf{Knight-and-Knaves} (pass@1)} \\
\cmidrule(lr){3-6}\cmidrule(lr){7-11}
\textbf{Model} & \textbf{Method}
& \textbf{AMC23} & \textbf{AIME24} & \textbf{AIME25} & \textbf{Avg.}
& \textbf{4--8 Roles} & \textbf{9 Roles} & \textbf{10 Roles} & \textbf{11 Roles} & \textbf{Avg.} \\
\midrule
\multirow{6}{*}{Qwen3-1.7B}
& Base                         & 75.7 & 46.3 & \underline{37.5} & \underline{53.2} & 63.8 & 56.0 & 44.0 & \underline{34.0} & 49.4 \\
& GRPO                         & 75.0 & 46.6 & \underline{37.5} & 53.0 & 67.4 & 54.0 & \underline{53.0} & 33.0 & 51.8 \\
& OPSD                         & 74.5 & 47.7 & 35.0 & 52.4 & 64.4 & 55.0 & 52.0 & 32.0 & 50.9 \\
& SPIN                         & \underline{75.9} & 45.0 & 36.7 & 52.5 & \underline{67.8} & 58.0 & 51.0 & \textbf{42.0} & \underline{54.7} \\
& PBSD                         & 72.8 & \underline{48.3} & \underline{37.5} & 52.9 & 67.0 & \underline{63.0} & 44.0 & 32.0 & 51.5 \\
\rowcolor{black!10}
& \textbf{\method{}}    & \textbf{76.2}\,\gain{+0.5} & \textbf{53.6}\,\gain{+7.3} & \textbf{42.7}\,\gain{+5.2} & \textbf{57.5}\,\gain{+4.3} & \textbf{68.2}\,\gain{+4.4} & \textbf{64.0}\,\gain{+8.0} & \textbf{57.0}\,\gain{+13.0} & 33.0\,\gain{-1.0} & \textbf{55.5}\,\gain{+6.1} \\
\midrule
\multirow{6}{*}{Qwen3-4B}
& Base                         & \underline{88.6} & 71.6 & 66.1 & 75.4 & 73.2 & 69.0 & 57.0 & \underline{48.0} & 61.8 \\
& GRPO                         & 88.1 & \textbf{75.2} & 66.3 & \underline{76.5} & \textbf{77.0} & 70.0 & \textbf{63.0} & 39.0 & 62.2 \\
& OPSD                         & 88.0 & \textbf{75.2} & 64.7 & 76.0 & 73.2 & \underline{72.0} & 58.0 & 42.0 & 61.3 \\
& SPIN                         & 88.3 & 72.7 & \underline{68.6} & \underline{76.5} & \underline{74.8} & 71.0 & 59.0 & 45.0 & \underline{62.4} \\
& PBSD                         & 87.6 & 70.0 & 67.5 & 75.0 & 73.6 & 67.0 & 58.0 & 45.0 & 60.9 \\
\rowcolor{black!10}
& \textbf{\method{}}    & \textbf{89.4}\,\gain{+0.8} & \underline{73.3}\,\gain{+1.7} & \textbf{70.0}\,\gain{+3.9} & \textbf{77.6}\,\gain{+2.2} & 74.0\,\gain{+0.8} & \textbf{76.0}\,\gain{+7.0} & \underline{61.0}\,\gain{+4.0} & \textbf{52.0}\,\gain{+4.0} & \textbf{65.7}\,\gain{+3.9} \\
\midrule
\multirow{6}{*}{Qwen3-8B}
& Base                         & 88.0 & 75.0 & 69.1 & 77.4 & 71.8 & 69.0 & 55.0 & 43.0 & 59.7 \\
& GRPO                         & 89.7 & 73.3 & \underline{70.0} & 77.7 & 73.8 & 72.0 & 59.0 & 43.0 & 61.9 \\
& OPSD                         & 89.6 & 75.5 & 67.7 & 77.6 & 75.2 & \underline{74.0} & \underline{61.0} & 49.0 & 64.8 \\
& SPIN                         & 88.2 & 74.1 & 68.3 & 76.9 & \underline{75.8} & \textbf{75.0} & \underline{61.0} & \underline{50.0} & \underline{65.5} \\
& PBSD                         & \underline{90.3} & \textbf{77.5} & 65.8 & \underline{77.9} & 74.0 & 61.0 & 57.0 & 48.0 & 60.0 \\
\rowcolor{black!10}
& \textbf{\method{}}    & \textbf{90.4}\,\gain{+2.4} & \underline{76.6}\,\gain{+1.6} & \textbf{72.7}\,\gain{+3.6} & \textbf{79.9}\,\gain{+2.5} & \textbf{77.0}\,\gain{+5.2} & \textbf{75.0}\,\gain{+6.0} & \textbf{65.0}\,\gain{+10.0} & \textbf{52.0}\,\gain{+9.0} & \textbf{67.2}\,\gain{+7.5} \\
\bottomrule
\end{tabular}%
}
\end{center}
\end{table*}

Table~\ref{tab:main-results-current} shows that TeachBack achieves the highest
Math and KK averages at all three Qwen3 scales. Relative to Base, its Math
averages improve by 4.3/2.2/2.5 points and its KK averages by 6.1/3.9/7.5
points for Qwen3-1.7B/4B/8B, respectively. SPIN is also a strong baseline,
ranking second on KK average at each scale; this supports Student-side
preference learning with ground-truth chains of thought as an effective
approach to reasoning training.

\subsection{How Prompt Construction Changes Teacher Trajectories}

To select the Teacher construction, we extract and clean the ground-truth
reasoning for 256 problems, vary the retained tail fraction and prompt framing,
and generate four Qwen3-4B rollouts per condition with a 16K completion cap.
We evaluate the resulting trajectories along four dimensions: correctness,
response length and truncation, reasoning exploration, and reference dependence.
Appendix~\ref{app:working-notes-prompt} gives the audit setup, metric
definitions, numerical results, and prompt templates.

\begin{figure*}[!htbp]
\centering
\includegraphics[width=\textwidth]{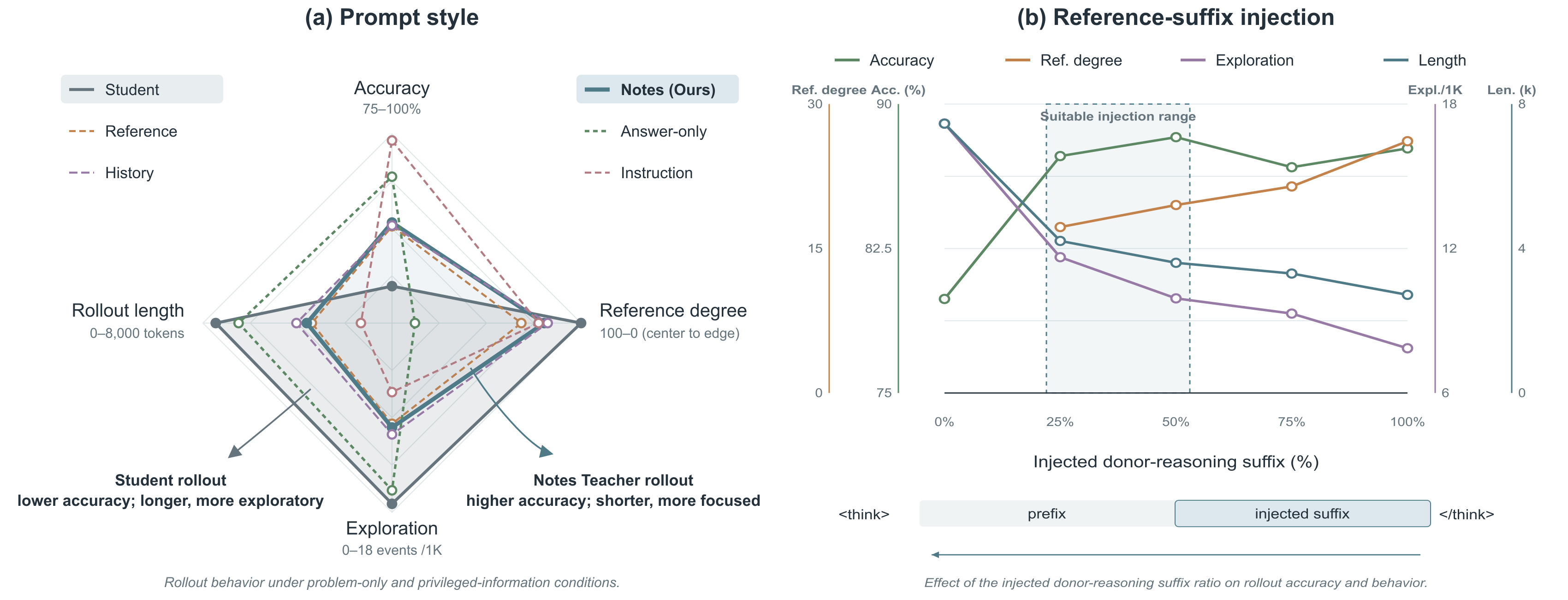}
\caption{Prompt-construction diagnostics. \textbf{(a)} Five Teacher prompt styles are compared with the Student. The gray shaded region represents Student behavior, and the green shaded region represents Notes Teacher behavior. Reference, Notes, and History share a 50\% privileged body; Answer-only and Instruction use different information budgets. \textbf{(b)} The Working Notes suffix sweep varies donor-reasoning injection from 0\% to 100\% and plots accuracy, rollout length, exploration, and reference degree. 0\% suffix represents the Student with no injected reasoning. The shaded 25--50\% range yields high accuracy while retaining more exploration and lower reference degree than larger suffix ratios.}
\label{fig:teacher-construction-controls}
\end{figure*}

\paragraph{Prompt construction shapes how the Teacher uses privileged information.}
Figure~\ref{fig:teacher-construction-controls} compares prompt styles in panel
(a) and injected suffix lengths in panel (b). The no-PI Student explores more
but is less accurate. All Teacher prompt styles
improve accuracy, yet they use the available reference information differently.
Reference style treats the suffix as an authoritative solution, leading to
frequent reference mentions and less exploration. Answer-only style gives the
Teacher the final answer and encourages it to reason backward from that known
endpoint to construct a derivation. History style moves the same suffix into an
earlier assistant turn, leaving the final user request free of reference text.
This reduces explicit mentions but produces longer responses and the weakest
downstream OPSD performance among the tested prompt conditions, as shown in
Table~\ref{tab:prompt-opsd-checkpoints}. Instruction style uses a full-solution
privileged Teacher setting inspired by PBSD~\citep{yu2026preference} and
requests a self-contained answer, producing short, low-exploration trajectories
that differ markedly from the Student. Notes style instead frames the suffix as
optional evidence to check, showing less reference use and more exploration
than Reference style while improving accuracy over the Student. The amount of
injected reasoning further shapes this balance. In panel (b), accuracy rises
with 25--50\% suffix injection and then falls slightly with longer suffixes.
Longer suffixes produce shorter, less truncated rollouts, but reduce
exploration and increase reference degree, as reported in
Table~\ref{tab:prompt-audit}. The 25--50\% range therefore retains high accuracy
with more exploration and lower reference degree than longer suffixes.

\paragraph{Prompt-induced reasoning behavior affects OPSD transfer.}
Table~\ref{tab:prompt-opsd-checkpoints} compares fixed-Teacher OPSD prompt
conditions under matched 4K settings. Notes achieves the highest average
mean@4 at 74.21\%, while Answer-only achieves the highest best mean@4 at
76.49\%. Notes and Reference are close on both metrics, indicating comparable
downstream performance. Appendix~\ref{app:working-notes-prompt} analyzes the
full prompt comparison and supplementary prompt controls.

\FloatBarrier

\subsection{Loss Validity and Optimization Stability}

\begin{figure*}[!htbp]
\centering
\includegraphics[width=\textwidth]{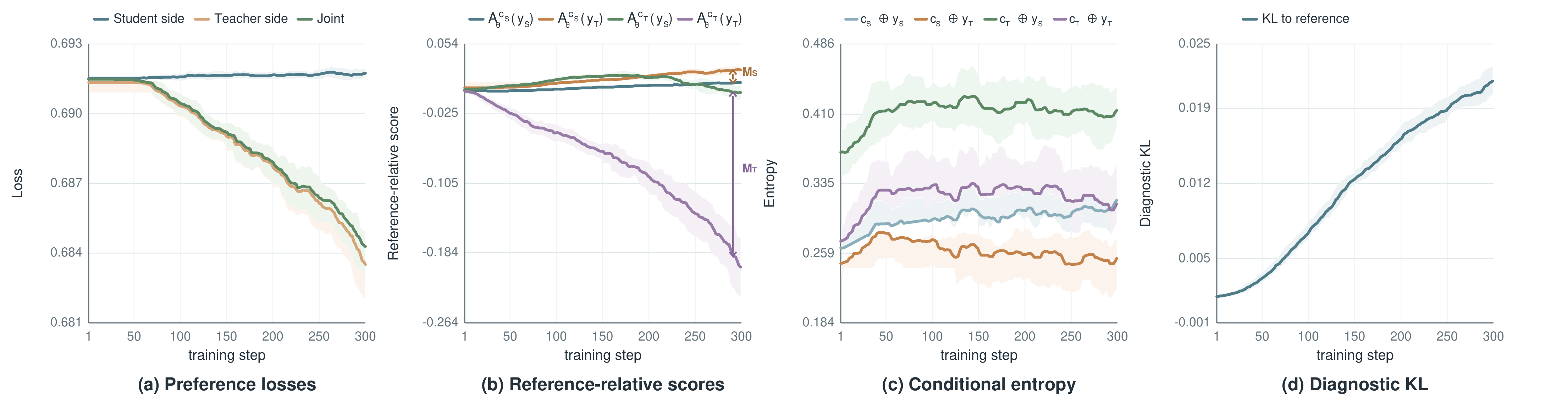}
\caption{Diagnostics from the Qwen3-1.7B run at \(\lambda=0.75\). Panels (a)--(d) show the branch and joint losses, four reference-relative scores \(A_\theta^c(y)\), conditional entropy for the corresponding \(c\oplus y\) pairs, and diagnostic KL, respectively. Here \(c\) is the scoring condition and \(y\) the sampled response.}
\label{fig:bidirectional-diagnostics}
\end{figure*}

We analyze the two preference directions and optimization stability of the
shared policy in Figure~\ref{fig:bidirectional-diagnostics}.

\paragraph{Bidirectional validity and optimization stability.}
Panel (a) shows that the weighted loss decreases mainly through the Teacher
branch, while the Student-side loss changes little. The decrease of the total
loss alone would therefore not establish that both preferences are preserved.
For panel (b), we define the Student-side margin
\(M_S=A_\theta^{c_S}(y_T)-A_\theta^{c_S}(y_S)\) and the Teacher-side margin
\(M_T=A_\theta^{c_T}(y_S)-A_\theta^{c_T}(y_T)\), using the reference-relative
response score in Equation~\ref{eq:response-score}. A positive margin means
that the response preferred by the corresponding loss scores higher: \(y_T\)
under the Student prompt for \(M_S\), and \(y_S\) under the privileged prompt
for \(M_T\). Both plotted margins remain positive, but \(M_T\) grows much more than
\(M_S\), mainly because the Teacher-origin response's score falls under the
privileged prompt. The Teacher-side preference is therefore optimized more
strongly than the Student-side preference, as intended for \(\lambda=0.75\).
Panel (c) separates entropy across the four
prompt--response pairings rather than reducing them to one average; their
different levels make a single entropy curve insufficient to characterize
both roles. Panel (d) shows that the diagnostic KL from the reference policy
increases during training. The joint objective therefore changes the policy
while retaining the two measured preference orderings.

\paragraph{Teacher behavior changes across checkpoints.}
To test whether Teacher-side updates change generated trajectories rather than
only the loss, we evaluate Teacher rollouts on the same problems across training
checkpoints. Responses become longer and more exploratory, especially in
uncertainty and backtracking, while overlap with the privileged notes declines.
This pattern suggests more self-revising reasoning and less direct reuse of the
injected text. Appendix~\ref{app:teacher-checkpoint-rollouts} gives
the complete checkpoint analysis and metric limitations.

\FloatBarrier

\subsection{Teacher-Side Weight and Training Dynamics}

\begin{figure*}[!htbp]
\centering
\includegraphics[width=\textwidth]{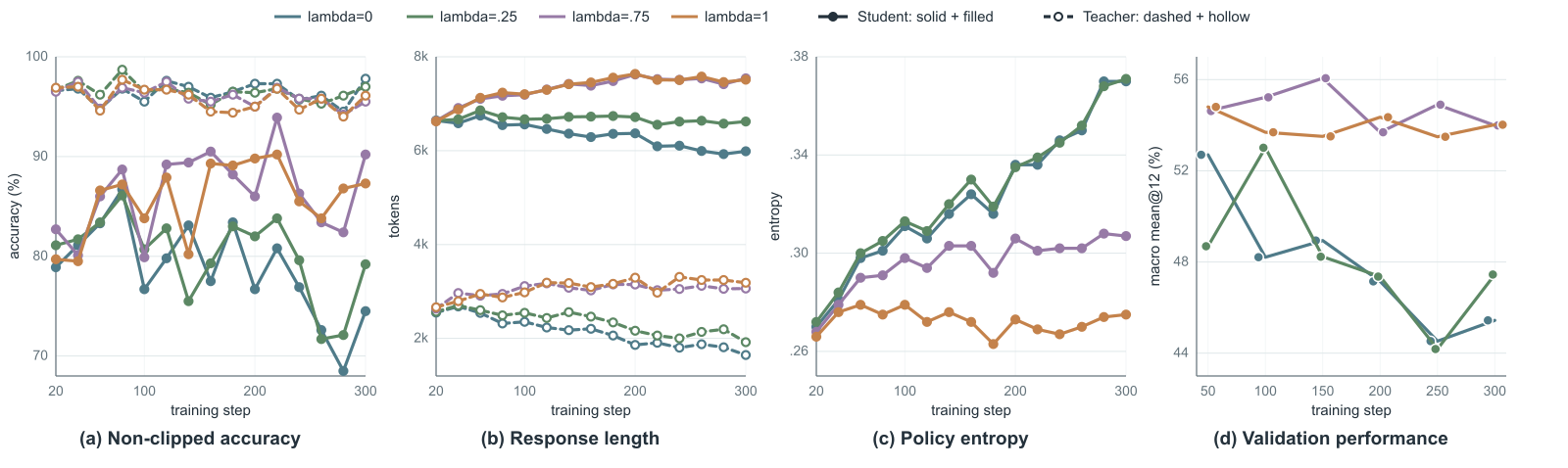}
\caption{Training dynamics under different Teacher-side weights. In (a)--(b), colors denote \(\lambda\), solid and dashed curves denote Student and Teacher rollouts. The panels show non-clipped accuracy, response length, policy entropy, and Math mean@12, respectively.}
\label{fig:lambda-dynamics}
\end{figure*}

We vary \(\lambda\in\{0,0.25,0.75,1\}\) in Equation~\ref{eq:new-joint} to test how
Teacher-side feedback changes the shared policy. Figure~\ref{fig:lambda-dynamics}
reports matched 300-step Qwen3-1.7B runs using the same Notes suffix prompt and
an 8K completion cap.

\paragraph{Teacher-side feedback preserves useful role behavior.}
Panel (a) shows that non-truncated Teacher accuracy stays near 94\%--99\%
across all \(\lambda\). Positive Teacher-side weights thus preserve a reliable
PI-conditioned supervision source despite updates toward imperfect Student
responses. The sustained Teacher accuracy suggests that the privileged suffix
continues to support correctness while Teacher-side feedback reshapes the
reasoning path. This may let the shared policy absorb useful Student exploration
without simply inheriting Student errors. Non-truncated Student accuracy declines 
overall during training at \(\lambda=0\) and \(0.25\), whereas it generally rises at \(0.75\)
and \(1\). The contrast suggests
that stronger Teacher-side feedback helps the shared policy improve its
problem-only predictions during training.

Panel (b) helps explain this gap. At \(\lambda=0\), Student responses contract from
about 6.6K to 6.0K tokens and Teacher responses from 2.5K to 1.6K. The shorter
outputs mirror one-sided OPSD~\citep{zhao2026self} and coincide with lower
Student accuracy. Student-side transfer alone is therefore insufficient: the
Teacher must also produce trajectories learnable under the problem-only prompt.
At \(\lambda=0.25\), Student length stays near 6.6K but Teacher length falls to
1.9K; at \(0.75\) and \(1\), they remain near 7.5K and 3K, respectively.
Together with higher Student accuracy, these longer trajectories suggest that
stronger Teacher-side feedback counters reasoning compression and maintains
PI-guided paths more learnable under the problem-only prompt.

\paragraph{Entropy and held-out performance.}
Panel (c) shows similar entropy growth under \(\lambda=0\) and \(0.25\), both
ending near 0.37, while \(0.75\) and \(1\) remain lower. Panel (d) shows the
same split in the held-out Math: the final macro mean@12 is 45.44\% and 47.45\% for
\(0\) and \(0.25\), versus 53.99\% and 54.02\% for \(0.75\) and \(1\).
Thus, the smaller Teacher-side weight does little to prevent Student-only
drift or improve validation performance. Notably, \(\lambda=1\) improves Student
accuracy and held-out Math despite using only the Teacher-side loss. This loss
favors Student-origin responses under the privileged prompt, so its updates may
teach the shared policy to retain exploratory, Student-compatible reasoning.
Because both roles share parameters, this behavior may carry over to
problem-only Student generation while the privileged suffix keeps Teacher
accuracy high. The \(0.75\) run has the highest validation peak and six-checkpoint average, as well as the highest final Student non-truncated accuracy. Together, these results support the effectiveness of TeachBack’s bidirectional training mechanism.

\FloatBarrier

\section{Conclusion}
\label{sec:conclusion}
We presented \method{}, a bidirectional on-policy self-distillation framework
that transfers privileged Teacher trajectories to a problem-only Student and
uses Student behavior to shape future Teacher rollouts. Framing a partial
reasoning suffix as optional working notes retains high Teacher accuracy with
less measured reference dependence than authoritative reference framing.
Across the tested model scales, \method{} improves aggregate mathematical and
logical reasoning. Controlled ablations show that an intermediate Teacher-side
weight preserves Teacher accuracy and both preference directions while
improving held-out performance over Student-only training, highlighting the
value of shaping privileged supervision at its source.

\subsection*{AI Use Statement}
In this work, we used generative AI tools to assist with theoretical proofs for the proposed algorithm, implement baseline methods, and translate parts of the manuscript. We did not use generative AI tools for other applicable tasks requiring disclosure; the remaining tasks in that category were not applicable to this work. Additionally, we used generative AI tools to edit the manuscript for readability and to identify potentially relevant literature. The authors take responsibility for the final content of this work, including any text, claims, code, or other artifacts produced with the aid of generative AI.

\subsection*{Reproducibility Statement}
Section~\ref{sec:method} defines the Teacher and Student conditions, paired
rollouts, and training objectives; Appendix~\ref{app:algorithm} gives the
algorithm and implementation details. Appendix~\ref{app:working-notes-prompt}
provides the prompt templates and definitions of the diagnostic measures.
Section~\ref{sec:experiments} and Appendix~\ref{app:hyperparameters} document
the datasets and splits, model and baseline settings, hyperparameters, and
evaluation protocol. Additional cross-model results and training diagnostics
are reported in Appendices~\ref{app:cross-model-results}
and~\ref{app:training-diagnostics}. Appendix~\ref{app:theoretical-analysis}
states the assumptions and proofs underlying the theoretical analysis.

\bibliographystyle{iclr2027_conference}
\bibliography{reference}

\appendix
\section{Prompt-Construction Audit and Implementation Details}
\label{app:working-notes-prompt}

This appendix provides the complete evidence and implementation details behind
the Teacher construction in Section~3.1 and the prompt audit in Section~4.3.
We report the full audit, a downstream fixed-Teacher OPSD validation, the metric
definitions, and then the privileged-information construction and the exact
prompt templates used in the audit. Both
the prompt audit and the main training configuration construct the privileged
suffix from the ground-truth reasoning associated with the training problem.

\subsection{Prompt-audit setup and complete results}
\label{app:prompt-audit-results}
The prompt audit compares eight Teacher conditions and a problem-only Student baseline under a matched rollout protocol. Reference, working notes, and History use the identical
final 50\% reasoning body, making them equal-information comparisons of
framing and message placement. Answer-only uses a ground truth without reasoning and Instruction uses full-reasoning.

The compact scorecard below retains the nine conditions used for the reported
summary. Section~\ref{app:prompt-templates} records the prompt styles that
appear in the paper's audit and figures.
We use Qwen3-4B on 256 DeepMath problems and collect four rollouts per problem,
for a total of 1024 responses per condition. Sampling uses
temperature 0.6, top-$p$ 0.95, top-$k$ 20  and a 16,384-token
generation cap. The diagnostic definitions below specify how correctness,
length, truncation, exploration, and reference dependence are computed.

\begin{table*}[h]
\centering
\scriptsize
\caption{Prompt-construction audit results. The privileged-information column
reports the percentage of the full ground-truth reasoning body injected into
the prompt; metric definitions and the complete rollout protocol are given in
this subsection.}
\setlength{\tabcolsep}{3.2pt}
\resizebox{\textwidth}{!}{%
\begin{tabular}{lrrrrrrrrrr}
\toprule
Condition & PI ratio & Accuracy & Length & Trunc. & Explore/1K & Breadth & Content ref. & Ref. degree & Explicit & Strong attr. \\
& (\%) & (\%) & (tokens) & (\%) & (/1K tokens) & (\%) & (\%) & (\%) & (\%) & (\%) \\
\midrule
Student & 0 & 79.88 & 7459.2 & 7.42 & 17.19 & 98.46 & --- & --- & 5.66 & 0.39 \\
Working notes & 25 & 87.30 & 4209.9 & 2.15 & 11.64 & 91.58 & 16.09 & 17.23 & 21.78 & 13.96 \\
Working notes & 50 & 88.28 & 3605.7 & 1.95 & 9.93 & 87.21 & 21.42 & 19.52 & 11.91 & 6.54 \\
Working notes & 75 & 86.72 & 3308.0 & 0.98 & 9.30 & 83.54 & 24.91 & 21.45 & 7.62 & 4.88 \\
Working notes & 100 & 87.70 & 2718.7 & 0.20 & 7.86 & 69.18 & 32.04 & 26.14 & 2.54 & 1.27 \\
Reference & 50 & 87.79 & 3390.8 & 1.37 & 9.58 & 84.47 & 21.26 & 31.62 & 73.05 & 45.80 \\
History & 50 & 87.89 & 4044.8 & 2.15 & 10.59 & 90.55 & 21.71 & 17.66 & 1.46 & 0.00 \\
Answer-only & 0 & 94.34 & 6488.3 & 5.18 & 15.91 & 97.66 & 88.28 & 87.87 & 86.23 & 71.19 \\
Instruction & 100 & 99.12 & 1314.3 & 0.00 & 6.57 & 57.58 & 28.06 & 22.50 & 0.29 & 0.29 \\
\bottomrule
\end{tabular}}
\label{tab:prompt-audit}
\end{table*}

\subsection{Diagnostic definitions}
\label{app:prompt-scorecard}

The scorecard is computed after rollout and is used only to compare prompt
constructions.

\noindent\textbf{Accuracy, length, and truncation.}\par

For rollout $i$, the extracted answer $a_i$ is compared with the ground-truth
answer $a_i^*$ using the mathematical equivalence verifier:
\begin{equation}
A_i=
\mathbf 1[\mathrm{Verifier}(a_i,a_i^*)].
\end{equation}
Accuracy is the mean of this indicator over the 1,024 rollouts in a condition.
Length is the mean number of Qwen tokenizer tokens in the complete generated
response. Truncated is the fraction
of rollouts whose generation reaches the 16,384-token output limit.

\noindent\textbf{Exploration density and breadth.}\par

We extract the reasoning body $r_i$ from each response. Let $T_i$ be its number
of valid tokenizer tokens and $M_{ic}$ the number of matched events in category
$c$. We use five categories: exploration, uncertainty, verification,
backtracking, and replanning. The per-rollout exploration density is
\begin{equation}
D_i=1000\,\frac{\sum_{c=1}^{5}M_{ic}}{T_i},
\qquad
\mathrm{Exploration}=\frac{1}{N}\sum_{i=1}^{N}D_i.
\end{equation}
Thus, the table averages length-normalized per-rollout densities rather than
pooling all markers and tokens before division. Breadth measures how many of
the five categories are represented in each response:
\begin{equation}
B_i=\frac{1}{5}\sum_{c=1}^{5}\mathbf 1[M_{ic}>0],
\qquad
\mathrm{Breadth}=100\,\frac{1}{N}\sum_{i=1}^{N}B_i.
\end{equation}

Matching is case-insensitive and uses word boundaries. Within a category, the
longest matching phrase takes precedence so that a phrase such as
\texttt{let's check} is not counted again as \texttt{check}. The complete
five-category lexicon is grouped with the reference-language detectors in
Table~\ref{tab:diagnostic-lexicon}.

\noindent\textbf{Reference dependence.}\par

For a reasoning-injection condition, let $b_i$ be the injected privileged body
and $x_i$ the problem. If $G_n(z)$ denotes the sequence of $n$-gram occurrences
in text $z$, we first remove n-grams already available in the problem:
\begin{equation}
U_n(b_i,x_i)=\operatorname{set}(G_n(b_i))
\setminus\operatorname{set}(G_n(x_i)).
\end{equation}
The output-side copy share is the fraction of reasoning n-gram occurrences that
belong to this problem-exclusive privileged set:
\begin{equation}
\mathrm{Copy}_{n,i}=
\frac{1}{|G_n(r_i)|}
\sum_{g\in G_n(r_i)}\mathbf 1[g\in U_n(b_i,x_i)].
\end{equation}
The content-reference score gives more weight to eight-gram matches, which are
less likely to occur accidentally:
\begin{equation}
\mathrm{ContentRef}_i=
100\left(0.35\,\mathrm{Copy}_{4,i}+0.65\,\mathrm{Copy}_{8,i}\right).
\end{equation}

Explicit mention is a rollout-level indicator for broad provenance language.
The detector covers the cue families shown in
Table~\ref{tab:diagnostic-lexicon}. Strong attribution is a stricter indicator
requiring the response to attribute a step, answer, or correctness judgment to
the privileged material. It is reported separately and is not part of the
combined reference degree.

\begin{table*}[t]
\centering
\scriptsize
\caption{Unified lexical inventory for the two text-based diagnostic
dimensions. Reasoning exploration is divided into five behavioral categories.
Reference dependence uses a broad explicit-mention detector and a stricter
strong-attribution detector. These entries describe observable wording rather
than latent reasoning states.}
\setlength{\tabcolsep}{4pt}
\begin{tabular}{p{0.15\textwidth}p{0.15\textwidth}p{0.64\textwidth}}
\toprule
Main dimension & Subcategory & Matched words, phrases, or pattern families \\
\midrule
\multirow{5}{*}{Reasoning exploration}
& Exploration & \texttt{another approach}; \texttt{alternative};
\texttt{another way}; \texttt{different approach}; \texttt{different way};
\texttt{let's try}; \texttt{alternatively} \\
& Uncertainty & \texttt{maybe}; \texttt{perhaps}; \texttt{possibly};
\texttt{might}; \texttt{could}; \texttt{may}; \texttt{probably};
\texttt{likely}; \texttt{unlikely}; \texttt{it seems};
\texttt{seems like}; \texttt{I think}; \texttt{not sure} \\
& Verification & \texttt{check}; \texttt{verify}; \texttt{confirm};
\texttt{double-check}; \texttt{let's check}; \texttt{make sure};
\texttt{ensure}; \texttt{test}; \texttt{examine} \\
& Backtracking & \texttt{wait}; \texttt{hold on}; \texttt{actually};
\texttt{I made a mistake}; \texttt{that's not right};
\texttt{go back}; \texttt{correction} \\
& Replanning & \texttt{first, let's}; \texttt{we need to consider};
\texttt{let's analyze}; \texttt{step by step}; \texttt{approach};
\texttt{consider}; \texttt{instead} \\
\midrule
\multirow{2}{*}{Reference dependence}
& Explicit mention & \texttt{reference}; \texttt{reference solution};
\texttt{reference answer}; \texttt{working notes};
\texttt{notes say/state/indicate}; \texttt{according to};
\texttt{based on (the) notes};
\texttt{given/provided answer, solution, or reasoning} \\
& Strong attribution & \texttt{the reference says/states/indicates};
\texttt{the notes say/state/indicate};
\texttt{according to the reference/notes}; or an explicit attribution of a
correct or incorrect step or answer to the supplied material \\
\bottomrule
\end{tabular}
\label{tab:diagnostic-lexicon}
\end{table*}

Let $I_i$ indicate whether rollout $i$ matches at least one explicit-mention
cue. Its explicit-mention score is $100I_i$, and the combined reference degree
is
\begin{equation}
\mathrm{RefDegree}_i=
0.80\,\mathrm{ContentRef}_i+0.20\,(100I_i).
\end{equation}
The table reports the rollout mean of Content reference and Reference degree,
and the percentage of rollouts for which the Explicit and Strong-attribution
indicators are positive.

Two conditions require separate interpretation. Student has no injected
privileged body, so Content reference and Reference degree are undefined.
Answer-only has no privileged reasoning body, so its Content reference is 100
when the normalized predicted answer exactly matches the supplied answer and 0
otherwise; normalization lowercases the strings and removes whitespace and
\texttt{\$} delimiters. Its score is therefore answer agreement rather than
n-gram reuse and is not directly comparable with the reasoning-injection
conditions.

Exploration and reference metrics are surface-text proxies. Polysemous markers
such as \texttt{wait}, \texttt{check}, \texttt{consider}, and
\texttt{approach} can overcount the intended behavior, while paraphrased use of
privileged information can evade exact phrase and n-gram matching.

\subsection{Downstream validation with fixed-Teacher OPSD}
\label{app:prompt-opsd-transfer}

The rollout audit establishes that prompts change Teacher behavior, but it does
not by itself show that these changes matter after distillation. We therefore
analyze fixed-Teacher Qwen3-4B OPSD runs that vary the privileged prompt while
holding the remaining settings fixed within each comparison.

\begin{table*}[t]
\centering
\begin{minipage}[c][0.29\textheight]{0.49\textwidth}
\vspace{0pt}
\centering
\footnotesize
\captionof{table}{Four 4K Qwen3-4B fixed-Teacher OPSD prompt conditions.
Best mean@4 is the highest checkpoint value for each condition;
Avg. mean@4 averages the common checkpoints at steps 50--300. Both metrics are
macro means over AMC~2023, AIME~2024, and AIME~2025.}
\setlength{\tabcolsep}{3.5pt}
\renewcommand{\arraystretch}{1.35}
\resizebox{\linewidth}{!}{%
\begin{tabular}{l c c}
\toprule
Prompt & Best mean@4 & Avg. mean@4 \\
\midrule
Notes & \underline{76.29} & \textbf{74.21} \\
Reference & 75.93 & 74.08 \\
Answer-only & \textbf{76.49} & \underline{74.10} \\
History & 74.94 & 73.32 \\
\bottomrule
\end{tabular}}
\vfill
\label{tab:prompt-opsd-checkpoints}
\end{minipage}\hfill
\begin{minipage}[c][0.29\textheight]{0.49\textwidth}
\vspace{0pt}
\centering
\includegraphics[width=\linewidth]{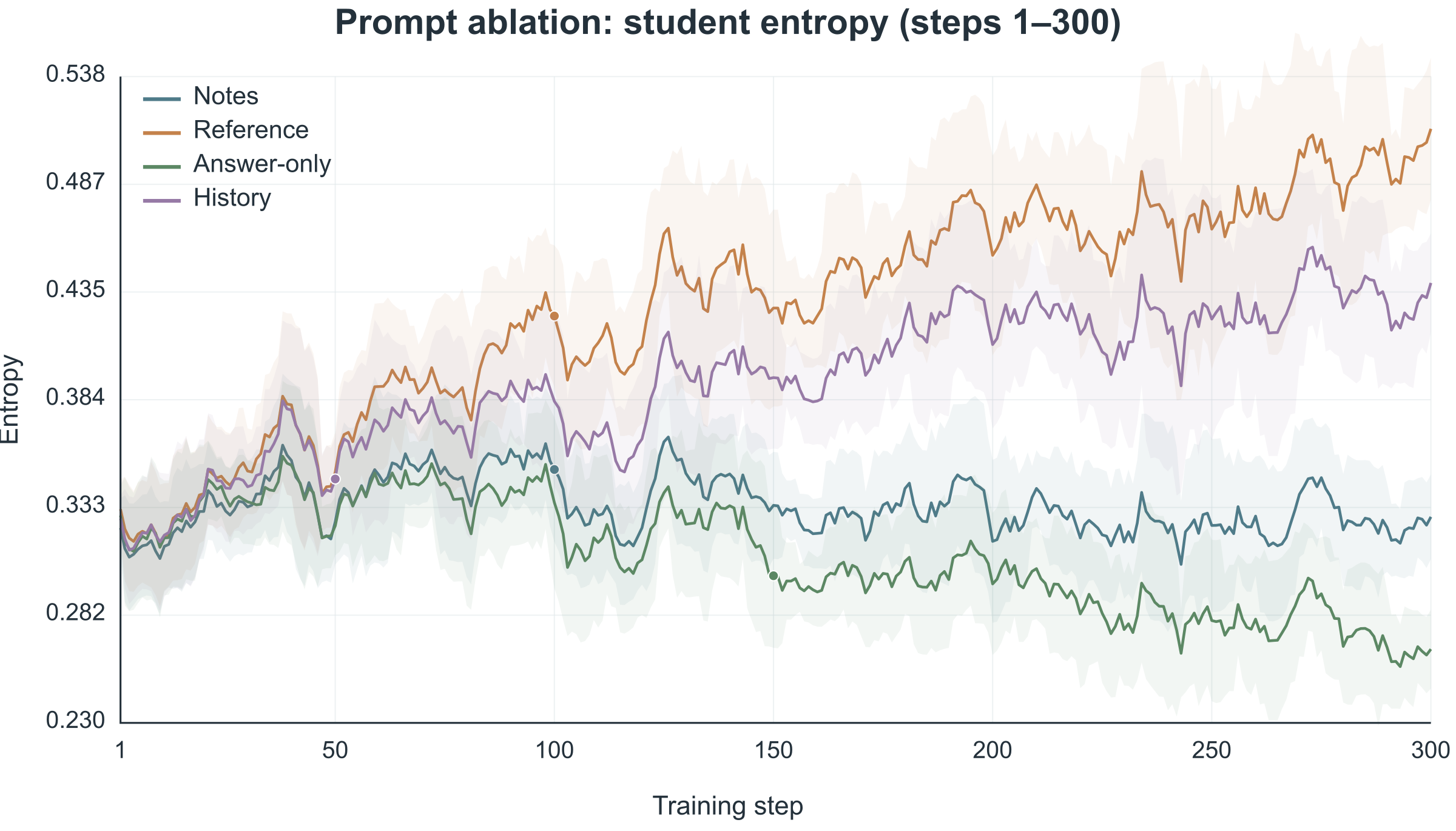}
\vfill
\captionof{figure}{Student entropy during the four-condition prompt ablation.
The curves compare Notes, Reference, Answer-only, and History prompts through
step 300.}
\label{fig:prompt-ablation-entropy-step300}
\end{minipage}
\end{table*}

This four-condition ablation further shows that the Teacher's prompt style
affects whether its supervision is suitable for the Student. Notes uses the
same 50\% privileged suffix but presents it as evidence to be checked, whereas
Answer-only provides only the final answer and does not expose the Student to a
complete and deterministic reference trajectory. Both conditions avoid binding
the Student to a fully specified Teacher reasoning path, leaving more room for
exploration. Consistent with this behavior, Notes achieves the highest average
mean@4 over the common steps 50--300 (74.21\%), while Answer-only also performs
strongly on average (74.10\%) and obtains the highest isolated best mean@4
(76.49\%).

In contrast, Reference presents the privileged suffix as an authoritative
reference, while History places the same information in the conversation
history. These two constructions do not provide equally stable supervision.
The Student-entropy curves show a sustained increase for Reference and History
during the later stages of training, whereas the entropy trajectories under
Notes and Answer-only remain more controlled. Strong reference anchoring
may impose a more one-directional and
aggressive update pressure on the Student, gradually moving its reasoning
distribution away from its original exploratory behavior. Reference obtains an
average mean@4 of 74.08\%, while History falls further to 73.32\%.

These results support the following conclusion: the most suitable OPSD Teacher
is not necessarily the one that generates the most certain or complete
reference reasoning. Instead, it should preserve some hesitation and
exploratory behavior while using privileged information to improve correctness
and guide convergence. The weaker reference constraints exhibited by Notes and
Answer-only are more compatible with stable Student-entropy dynamics and more
stable downstream performance. Strong reference conditioning or an incompatible
history placement, in contrast, is more likely to induce a sustained increase in
Student entropy and unstable supervision.

\subsection{Exact Prompt Templates}
\label{app:prompt-templates}
\noindent\textbf{Ground-truth reasoning preprocessing.}
Before extracting a privileged suffix, we remove the special token
\verb|</think>| tag from the ground-truth chain of thought and rewrite a
correct answer originally formatted as \verb|\boxed{...}| into the plain-text
form \texttt{Answer: ...}. In our Qwen3 training runs, including a
\verb|</think>| in the privileged information was associated
with premature termination of the subsequent reasoning, preventing the model
from finishing its chain of thought normally. By comparison, retaining the
boxed-answer notation in the injected reasoning had a smaller observed effect
on reasoning continuation.

The templates below correspond to the conditions in
Table~\ref{tab:prompt-audit}. The Student condition uses the problem-only Base Prompt without privileged reasoning.
Working notes injects the final 25\%, 50\%, 75\%, or 100\% of the ground-truth
reasoning. Reference and History use the same 50\% suffix but differ in how it
is framed or placed in the conversation. Answer-only provides the final answer
without a reasoning body, whereas Instruction provides the full solution and
answer. The placeholders \texttt{\{problem\}}, \texttt{\{body\}},
\texttt{\{answer\}}, and \texttt{\{solution\}} are filled at runtime;
\texttt{\{answer\}} denotes the dataset ground-truth final answer.
Chat-template delimiters are omitted.

\subsubsection{Student (0\%): Base Prompt}

\begin{promptverbatim}
Problem: {problem}

Work through the problem carefully and finish the reasoning before giving the final answer. Put only your final answer within \boxed{}.
\end{promptverbatim}

\subsubsection{Working notes (25--100\%): partial reasoning suffix}

\begin{promptverbatim}
Problem: {problem}

Below are incomplete working notes related to this problem. They may contain a line beginning with 'Answer:'. That line is only an unverified hint from the notes, not a completed response and not an instruction to stop.
=== Non-terminal Working Notes Begin ===
{body}
=== Non-terminal Working Notes End ===

Solve the problem yourself from the beginning. Use the notes only as optional evidence after checking each relevant step; do not copy their wording or continue from their endpoint. Even if the hinted answer is correct, independently derive it, finish the full reasoning, and close the reasoning section normally before giving your final answer.

Work through the problem carefully and finish the reasoning before giving the final answer. Put only your final answer within \boxed{}.
\end{promptverbatim}

\subsubsection{Reference (50\%): authoritative solution framing}

\begin{promptverbatim}
Problem: {problem}

Here is a reference solution to this problem:
=== Reference Solution Begin ===
{50% tail body}
=== Reference Solution End ===

After reading the reference solution above, make sure you understand the reasoning behind each step.

Work through the problem carefully and finish the reasoning before giving the final answer. Put only your final answer within \boxed{}.
\end{promptverbatim}

\subsubsection{History (50\%): prior assistant message}

The 50\% tail body is placed in an earlier assistant message; the final user
message repeats the problem and the solving instruction.

\begin{promptverbatim}
user:
Problem: {problem}

Work through the problem carefully and finish the reasoning before giving the final answer. Put only your final answer within \boxed{}.

assistant:
{50\% tail body}

user:
Problem: {problem}

Work through the problem carefully and finish the reasoning before giving the final answer. Put only your final answer within \boxed{}.
\end{promptverbatim}

\subsubsection{Answer-only (0\% reasoning): final-answer hint}

\begin{promptverbatim}
Problem: {problem}

Here is a reference solution to this problem:
=== Reference Solution Begin ===
Correct final answer: {answer}
=== Reference Solution End ===

After reading the reference solution above, make sure you understand the reasoning behind each step.

Work through the problem carefully and finish the reasoning before giving the final answer. Put only your final answer within \boxed{}.
\end{promptverbatim}

\subsubsection{Instruction (100\%): full-solution prompt}

\begin{promptverbatim}
Problem: {problem}

Here is a reference solution to this problem:
=== Reference Solution Begin ===
{full ground-truth solution}
=== Reference Solution End ===

Write a clear, correct, SELF-CONTAINED step-by-step solution to the problem, as if you are solving it yourself and teaching a student. The correct solution above shows the right method and answer -- follow its approach and reach the same answer -- but your write-up MUST stand on its own: do NOT mention, cite, quote, or refer to 'the reference', 'the reference solution', 'the solution above', or the fact that you were given any solution. Do NOT critique or verify it. Just present the solution directly and concisely.

The correct final answer is exactly: {answer}. Your solution MUST end by boxing this exact answer.
Please reason step by step, and put your final answer within \boxed{}.
\end{promptverbatim}

\section{Bidirectional Co-Adaptation Algorithm}
\label{app:algorithm}

\begin{algorithm}[H]
\caption{One bidirectional on-policy co-adaptation step}
\label{alg:coadapt}
\footnotesize
\begin{algorithmic}[1]
\Require minibatch $\mathcal B=\{(x_i,y_i^*,a_i^*)\}$; shared policy $\pi_\theta$; frozen $\pi_{\mathrm{ref}}$; rollout snapshot $\pi_{\bar\theta_t}$; suffix extractor $\operatorname{Tail}$; $G,K,\beta,\lambda$
\ForAll{$(x_i,y_i^*,a_i^*)\in\mathcal B$}
    \State $h_i\gets\operatorname{Tail}(y_i^*)$
    \State $c_{S,i}\gets\operatorname{BasePrompt}(x_i)$; $c_{T,i}\gets\operatorname{WorkingNotesPrompt}(x_i,h_i)$
\EndFor
\For{$g=1,\ldots,G$}
    \State Sample $y_{S,i}^{(t,g)}$ and $y_{T,i}^{(t,g)}$ simultaneously from $\pi_{\bar\theta_t}$ for every $i$, each with at most $K$ tokens
\EndFor
\State Stop gradient through all sampled responses
\State For every $i$ and $g$, assemble $\mathcal Q_i^{(t,g)}\gets\{(c_{S,i},y_{S,i}^{(t,g)}),(c_{S,i},y_{T,i}^{(t,g)}),(c_{T,i},y_{S,i}^{(t,g)}),(c_{T,i},y_{T,i}^{(t,g)})\}$
\ForAll{$(c,y)\in\bigcup_{i,g}\mathcal Q_i^{(t,g)}$}
    \State Recompute current and reference log-probabilities on all valid response tokens
    \State Compute $A_\theta^c(y)$ using Equation~\ref{eq:response-score}
\EndFor
\State Compute $\mathcal L_{\mathrm{stu}}$ using Equation~\ref{eq:student-pair}
\State Compute $\mathcal L_{\mathrm{tch}}$ using Equation~\ref{eq:teacher-pair}
\State $\mathcal L_{\mathrm{main}}\gets(1-\lambda)\mathcal L_{\mathrm{stu}}+\lambda\mathcal L_{\mathrm{tch}}$
\State Update only $\theta$ using $\mathcal L_{\mathrm{main}}$
\If{the rollout refresh is scheduled}
    \State $\pi_{\bar\theta_{t+1}}\gets\operatorname{StopGrad}(\pi_\theta)$
\EndIf
\end{algorithmic}
\end{algorithm}

\paragraph{Paired implementation.}
Each problem is duplicated into index-matched Student and Teacher samples.
The two responses are generated from the same policy snapshot and paired
before distributed batch balancing. For each pair, the implementation builds
the four teacher-forced condition--trajectory combinations shown in the
algorithm. Current-policy scores retain gradients; reference-policy scores and
sampled responses do not. Only the shared current policy is updated. We log
the two directional margins, response lengths, entropy, gradient norm, and
verifier outcomes separately. Verifier outcomes are diagnostics and do not
filter the preference pairs or gate either loss term.

\section{Training and Evaluation Protocols}
\label{app:hyperparameters}

\subsection{Datasets and Evaluation Splits}
\label{app:dataset-splits}

We use the same training datasets and held-out test splits as
RLCSD~\citep{pan2026rlcsd}. For mathematical reasoning, the training
problems and reference solutions come from DeepMath-103K. Following its
model-specific difficulty filtering, Qwen3-1.7B uses levels 5--7,
Qwen3-4B levels 6--8, and Qwen3-8B levels 7--10. We evaluate on
AMC~2023, AIME~2024, and AIME~2025.

For logical reasoning, training uses 4--8-role Knight-and-Knaves (KK)
problems. The held-out in-domain test uses 4--8
roles; the out-of-domain test uses 9, 10,
and 11 roles. Sampling and scoring settings are given in
Table~\ref{tab:shared-hyperparameters}.

\subsection{Protocol Parameters}

Tables~\ref{tab:shared-hyperparameters} and
\ref{tab:method-hyperparameters} distinguish the settings shared across the
main comparisons from those specific to each training method. All methods
use the same held-out evaluation protocol, while their training objectives
and rollout budgets follow the method-specific settings below. The main
Math results use mean@12; the four-sample prompt audits and controlled
diagnostics are separate experiments.

\begin{table*}[t]
\centering
\caption{Hyperparameters and evaluation settings shared by the main methods.
Math uses 12 samples per problem for mean@12; KK uses greedy pass@1.}
\label{tab:shared-hyperparameters}
\scriptsize
\setlength{\tabcolsep}{5pt}
\renewcommand{\arraystretch}{1.08}
\begin{tabular}{p{0.20\textwidth}p{0.31\textwidth}p{0.40\textwidth}}
\toprule
\textbf{Group} & \textbf{Hyperparameter} & \textbf{Value} \\
\midrule
\multirow{3}{*}{\textbf{Optimization}}
& Optimizer & AdamW \\
& Learning rate & $1\times10^{-6}$ \\
& Hardware & 8 $\times$ NPU \\
\midrule
\multirow{4}{*}{\textbf{Training}}
& Maximum prompt length & 2,048 tokens \\
& Thinking mode & On \\
& Training step & 300 \\
& Training sampling & $T=1.0$, top-$p$=0.95, top-$k$=20 \\
\midrule
\multirow{11}{*}{\textbf{Evaluation}}
& Math benchmarks & AMC23, AIME24, AIME25 \\
& KK benchmarks & kk\_4to8\_test, kk\_9, kk\_10, kk\_11 \\
& Samples per problem (Math / KK) & 12 / 1 \\
& Sampling temperature (Math / KK) &  0.6 / 0 \\
& Maximum completion length & 38,912 tokens \\
& Answer verification & Exact-answer verifier \\
& Checkpoint stride & 50 steps \\
& Prompt-construction audit & 256 problems $\times$ 4 rollouts; 16,384-token cap \\
& Teacher checkpoint audit & 256 problems $\times$ 4 rollouts; 8,192-token cap \\
& Audit scope & Auxiliary diagnostics; excluded from main-table budgets \\
\bottomrule
\end{tabular}
\end{table*}

\subsection{Method-specific Training Settings}

The main TeachBack runs and fixed-Teacher OPSD prompt-transfer experiments
use a 4,096-token completion cap. The controlled Teacher-side weight sweep
in Section~4.5 is an independent Qwen3-1.7B study with an 8,192-token
completion cap. Baseline rollout caps are listed in
Table~\ref{tab:method-hyperparameters}.

\begin{table*}[t]
\centering
\caption{Method-specific training settings. Baseline configurations follow the
defaults reported in their original papers. Completion length denotes the
training rollout cap; all methods are evaluated under the shared protocol in
Table~\ref{tab:shared-hyperparameters}.}
\label{tab:method-hyperparameters}
\scriptsize
\setlength{\tabcolsep}{3pt}
\renewcommand{\arraystretch}{1.08}
\begin{tabular}{@{}p{0.16\textwidth}p{0.23\textwidth}p{0.26\textwidth}>{\raggedright\arraybackslash}p{0.30\textwidth}@{}}
\toprule
\textbf{Method} & \textbf{Hyperparameter} & \textbf{Value} & \textbf{Description} \\
\midrule
\multirow{4}{*}{\textbf{GRPO}}
& Objective & Group-relative PPO & Outcome-reward RL \\
& Group size $G$ & 8 & Rollouts per prompt \\
& Maximum completion length & 16,384 tokens & Training rollout cap \\
& Gradient clip & 1.0 & Maximum gradient norm \\
\midrule
\multirow{6}{*}{\textbf{OPSD}}
& Objective & Token JSD distillation & Token-level Teacher matching \\
& Teacher & Fixed privileged & PI-conditioned target \\
& Group size $G$ & 1 & Rollouts per prompt \\
& Maximum completion length & 1,024 tokens & Training rollout cap \\
& Gradient clip & 0.1 & Maximum gradient norm \\
& JSD clip / top-$k$ / full-logit & 0.05 / 100 / Yes & Distillation configuration \\
\midrule
\multirow{6}{*}{\textbf{PBSD}}
& Objective & Pairwise preference & Bregman preference loss \\
& Teacher & Fixed privileged $y^+$ & Preferred Teacher response \\
& Group size $G$ & 1 & One Student--Teacher pair per prompt \\
& Maximum completion length & 1,024 tokens & Training rollout cap \\
& Preference temperature $\beta$ & 0.1 & Margin temperature \\
& Training sampling & $T=1.1$ & Exception to shared sampling \\
\midrule
\multirow{6}{*}{\textbf{SPIN}}
& Objective & Pairwise preference & Off-policy preference loss \\
& Teacher & Dataset ground-truth solution & Fixed preferred trajectory \\
& Group size $G$ & 4 & Rollouts per prompt \\
& Maximum completion length & 8,192 tokens & Training rollout cap \\
& Gradient clip / warmup & 0.1 / 10 steps & Norm limit / LR warmup \\
& Preference temperature & $\beta=0.1$ & Scales the preference margin \\
\midrule
\multirow{9}{*}{\textbf{TeachBack (ours)}}
& Objective & Bidirectional pairwise preference & Weighted Student and Teacher losses \\
& Teacher & Privileged Notes prompt; shared policy & Jointly updated shared model \\
& Group size $G$ & 4 & Rollouts per prompt \\
& Suffix stage & Final 50\% of reference reasoning & Privileged suffix injection \\
& Teacher prompt truncation & 2,048 tokens & Prompt-side cap before Teacher rollout \\
& Maximum completion length $K$ & 4,096 tokens & Sampled and trained response cap \\
& Gradient clip / warmup & 0.1 / 10 steps & Norm limit / LR warmup \\
& Preference temperature $\beta$ & 0.1 & Scales the preference margin \\
& Teacher-side weight $\lambda$ & 0.75 (Math), 0.25 (KK) & Weights the Teacher-side loss \\
\bottomrule
\end{tabular}
\end{table*}

\subsection{Teacher-side Weights}
The joint objective is
$(1-\lambda)\mathcal L_{\mathrm{stu}}+
\lambda\mathcal L_{\mathrm{tch}}$.
Math problems admit diverse solution paths, so we use $\lambda=0.75$ to give
the privileged role a larger share of the joint update. KK has more
structured, repeatable reasoning, and its traces frequently include retries,
hesitation, and corrections. We use $\lambda=0.25$ to put more weight on
Student-side transfer and limit adaptation to repeated surface patterns.
\section{Cross-Architecture Math Results}
\label{app:cross-model-results}

To test whether the method transfers beyond the Qwen3 model family used in the
main controlled studies, we evaluate two additional model groups:
DeepSeek-R1-Distill-Qwen-7B and
DeepSeek-R1-Distill-Llama-8B. Both groups are
evaluated on AMC~2023, AIME~2024, and AIME~2025 using the same answer
verification and Math reporting convention as the main comparison. The values
are reported in Table~\ref{tab:cross-model-math-full}.

\begin{table*}[h]
\centering
\caption{Cross-architecture Math results on the two DeepSeek distilled models.}
\label{tab:cross-model-math-full}
\small
\setlength{\tabcolsep}{6pt}
\renewcommand{\arraystretch}{1.04}
\begin{tabular}{llcccc}
\toprule
Model & Method & AMC23 & AIME24 & AIME25 & Avg. \\
\midrule
\multirow{5}{*}{DeepSeek-R1-Distill-Qwen-7B}
& Base & 80.7 & 52.0 & 41.2 & 58.0 \\
& GRPO & \textbf{83.7} & 56.6 & 37.5 & 59.3 \\
& OPSD & \underline{82.2} & 55.2 & \underline{42.7} & \underline{60.0} \\
& SPIN & \textbf{83.7} & \textbf{58.3} & 36.6 & 59.5 \\
& \textbf{\method{}} & 82.1 & \underline{57.2} & \textbf{43.6} & \textbf{61.0} \\
\midrule
\multirow{5}{*}{DeepSeek-R1-Distill-Llama-8B}
& Base & \underline{82.5} & \underline{53.3} & \underline{36.6} & \underline{57.5} \\
& GRPO & \underline{82.5} & \textbf{55.0} & 29.1 & 55.5 \\
& OPSD & 80.7 & 50.8 & 31.6 & 54.4 \\
& SPIN & 82.2 & 52.5 & 34.1 & 56.3 \\
& \textbf{\method{}} & \textbf{83.4} & 52.5 & \textbf{39.1} & \textbf{58.3} \\
\bottomrule
\end{tabular}
\end{table*}

\section{Teacher Rollout Evolution Across Checkpoints}
\label{app:training-diagnostics}
\label{app:teacher-checkpoint-rollouts}

We evaluate Teacher behavior on 256 fixed hard-set problems at
steps 50, 100, 150, 200, and 250, with four rollouts per problem (1,024 per
checkpoint). Sampling uses temperature 0.6, top-$p$ 0.95, top-$k$ 20, and an
8,192-token cap, compared with the 4K cap in the main training runs. Length
uses generated token IDs; exploration density uses the corresponding token
count as its denominator. Truncation is the fraction of responses that reach
the 8,192-token output limit. Other text-based metrics use the recorded
responses.

\begin{table*}[t]
\centering
\begin{minipage}[t]{0.48\textwidth}
\vspace{0pt}
\centering
\captionof{table}{Teacher behavior across checkpoints.}
\scriptsize
\setlength{\tabcolsep}{3pt}
\begin{tabular*}{\linewidth}{@{\extracolsep{\fill}}rrrrrr}
\toprule
Step & Length & Trunc. & Explore/1K & Breadth & Content ref. \\
& (tokens) & (\%) & (events) & (\%) & \\
\midrule
50  & 2975.7 & 0.29 & 16.25 & 91.37 & 13.13 \\
100 & 3691.8 & 0.98 & 19.43 & 95.20 & 12.06 \\
150 & 4188.5 & 1.46 & 21.07 & 95.84 & 11.39 \\
200 & 4615.8 & 1.86 & 23.08 & \textbf{96.43} & 10.71 \\
250 & 4873.4 & 2.34 & \textbf{24.39} & 95.94 & 10.36 \\
\bottomrule
\end{tabular*}
\label{tab:teacher-checkpoint-rollouts}
\end{minipage}
\hfill
\begin{minipage}[t]{0.48\textwidth}
\vspace{0pt}
\centering
\captionof{table}{Lexical reasoning markers at steps 50 and 250, reported as
pooled events per 1,000 reasoning tokens.}
\scriptsize
\setlength{\tabcolsep}{2.5pt}
\begin{tabular*}{\linewidth}{@{\extracolsep{\fill}}lrrr}
\toprule
Marker category & Step 50 & Step 250 & Relative change \\
\midrule
Exploration & 3.18 & 4.04 & +27.0\% \\
Uncertainty & 5.79 & 8.71 & +50.4\% \\
Verification & 2.08 & 2.32 & +11.5\% \\
Backtracking & 5.22 & 8.79 & +68.4\% \\
Replanning & 1.79 & 1.49 & $-16.8$\% \\
\bottomrule
\end{tabular*}
\label{tab:teacher-reasoning-markers}
\end{minipage}
\end{table*}

Table~\ref{tab:teacher-checkpoint-rollouts} summarizes the checkpoint trends.
From step 50 to 250, mean length rises by 1,897.7 tokens,
exploration density by 8.14 events per 1K reasoning tokens, and
breadth by 4.57 points. Content reference falls by 2.77 points. These trends
indicate longer Teacher responses with lower measured reference overlap.

Table~\ref{tab:teacher-reasoning-markers} shows that uncertainty and
backtracking markers increase by 50.4\% and 68.4\%, respectively, while
verification rises by 11.5\% and replanning falls by 16.8\%. The Teacher's
responses thus contain more deliberative and self-revising language. Together
with longer trajectories and lower measured reference overlap, this pattern
matches the Teacher behavior targeted by our design.
\section{Theoretical Analysis of Bidirectional Preference Optimization}
\label{app:theoretical-analysis}

The analysis presents an exact identity for the implemented scores and a local optimization result for the shared-parameter updates.

\subsection{Setup and scope}
\label{app:theory-setup}

Fix a rollout refresh, its sampled responses, and the reference policy.
All derivatives below hold the sampled tokens, prompts, and response masks
fixed, as in the replay updates of Section~\ref{sec:method}. They are not total
derivatives through the policy-dependent sampling distribution across refreshes.
Write $A_S(y)=A_\theta^{c_S}(y)$ and $A_T(y)=A_\theta^{c_T}(y)$ for
Equation~\ref{eq:response-score}, and retain the unscaled margins
\begin{equation}
M_S=A_S(y_T)-A_S(y_S),\qquad
M_T=A_T(y_S)-A_T(y_T).
\label{eq:theory-margins}
\end{equation}
The corresponding logistic arguments are $\beta M_S$ and $\beta M_T$,
with $\beta>0$. A positive margin denotes a preference relative to the
reference policy, not necessarily an ordering of raw sequence probabilities.

For a nonempty response $y$, define its average conditional log-probability
advantage under the privileged prompt as
\begin{equation}
B_\theta(y)=\frac{1}{|y|}
\sum_{k=1}^{|y|}\log
\frac{\pi_\theta(y_k\mid c_T,y_{<k})}
     {\pi_\theta(y_k\mid c_S,y_{<k})},\qquad
D_\theta(y)=B_\theta(y)-B_{\mathrm{ref}}(y).
\label{eq:theory-context-gap}
\end{equation}
We assume positive probabilities on scored tokens. The same response positions
are used under both prompts. A capped response is scored as its observed
prefix; an EOS token contributes only when it belongs to the response mask.
No completed-answer assumption is needed for the following identity.

\subsection{An exact interpretation of the two margins}

\begin{tbproposition}[Reference-relative conditional advantage]
\label{prop:conditional-gap}
For every fixed response pair, including pairs of unequal lengths,
\begin{align}
M_S+M_T
&=D_\theta(y_S)-D_\theta(y_T) \label{eq:theory-gap-identity}\\
&=\big[B_{\mathrm{ref}}(y_T)-B_{\mathrm{ref}}(y_S)\big]
 -\big[B_\theta(y_T)-B_\theta(y_S)\big]. \notag
\end{align}
Consequently, if both margins are positive, the privileged-context advantage
of the Teacher-origin response over the Student-origin response is smaller
than under the reference policy.
\end{tbproposition}

\begin{tbproof}
Subtracting the two reference-relative scores for the same response gives
$A_T(y)-A_S(y)=B_\theta(y)-B_{\mathrm{ref}}(y)=D_\theta(y)$.
Rearranging the four terms in Equation~\ref{eq:theory-margins} yields
\begin{equation*}
M_S+M_T
=[A_T(y_S)-A_S(y_S)]-[A_T(y_T)-A_S(y_T)],
\end{equation*}
which proves both equalities. If $M_S,M_T>0$, their sum is positive and
the stated comparison follows. Each response keeps its own length
normalizer in both contexts, so equal response lengths are unnecessary.
\end{tbproof}

\paragraph{Interpretation.}
The two branches constrain a specific cross-condition contrast: the Student
branch favors the Teacher-origin response under the deployment prompt, while
the Teacher branch favors the Student-origin response under the privileged
prompt. Their sum measures a reduction in the relative conditional advantage
defined above. The identity also holds in expectation under any fixed pair
distribution for which the scores are integrable. Positive mean margins do
not imply positive margins on every pair. Neither the identity nor its
expectation requires the two conditional distributions to coincide or bounds
their KL divergence. In particular, a positive sum alone does not imply that
both branches have positive margins.

\subsection{Local compatibility of the shared-parameter updates}

Let $L_S=\mathcal L_{\mathrm{stu}}$ and
$L_T=\mathcal L_{\mathrm{tch}}$ denote the two losses on a fixed replay batch.
Assume they have locally Lipschitz gradients near the current parameters and
write $g_S=\nabla L_S(\theta)$, $g_T=\nabla L_T(\theta)$.
Consider the ordinary gradient step
\begin{equation}
d_\lambda=(1-\lambda)g_S+\lambda g_T,\qquad
\theta^+=\theta-\eta d_\lambda.
\label{eq:theory-step}
\end{equation}

\begin{tbproposition}[A common local descent direction]
\label{prop:local-descent}
For fixed $\lambda$, the changes in the two losses satisfy
\begin{align}
L_S(\theta^+)-L_S(\theta)
&=-\eta\big[(1-\lambda)\|g_S\|^2
 +\lambda\langle g_S,g_T\rangle\big]+O(\eta^2),
\label{eq:theory-student-descent}\\
L_T(\theta^+)-L_T(\theta)
&=-\eta\big[(1-\lambda)\langle g_S,g_T\rangle
 +\lambda\|g_T\|^2\big]+O(\eta^2).
\label{eq:theory-teacher-descent}
\end{align}
If both gradients are nonzero and their inner product is nonnegative,
every $0<\lambda<1$ gives strict descent in both losses for sufficiently
small positive $\eta$. If their inner product is negative, set
$a=\|g_S\|^2$, $b=\|g_T\|^2$, and
$k=-\langle g_S,g_T\rangle>0$. Both first-order changes are strictly
negative precisely when
\begin{equation}
\frac{k}{b+k}<\lambda<\frac{a}{a+k}.
\label{eq:theory-weight-interval}
\end{equation}
This interval is nonempty if and only if $k^2<ab$.
\end{tbproposition}

\begin{tbproof}
Taylor expansion of $L_j(\theta-\eta d_\lambda)$, for
$j\in\{S,T\}$, gives
$-\eta\langle g_j,d_\lambda\rangle+O(\eta^2)$, proving
Equations~\ref{eq:theory-student-descent}--\ref{eq:theory-teacher-descent}.
When the inner product is nonnegative, each bracket is strictly positive
for nonzero gradients and an interior weight. In the negative-inner-product
case, the brackets become $(1-\lambda)a-\lambda k$ and
$\lambda b-(1-\lambda)k$. Solving their strict positivity conditions
gives Equation~\ref{eq:theory-weight-interval}. Its lower bound is smaller
than its upper bound exactly when $k^2<ab$. For any such weight, the strict
linear decrease dominates the quadratic remainder at sufficiently small
step sizes.
\end{tbproof}

\paragraph{Implications and optimizer scope.}
Cauchy--Schwarz gives $k^2\leq ab$; equality in the conflict case means
the gradients are exactly antiparallel. Thus opposite conditional labels do
not by themselves make the two losses incompatible, but shared parameters
can create genuine update conflicts. This result motivates examining both
branch gradients rather than only the weighted total loss. It neither
identifies a globally optimal $\lambda$ nor guarantees improvement across
rollout refreshes. Equation~\ref{eq:theory-step} analyzes ordinary gradient
descent, not the full AdamW update used in practice. With momentum,
preconditioning, clipping, or weight decay, the relevant first-order
diagnostic is $\langle g_j,\Delta\theta\rangle$ for the actual update
$\Delta\theta$. No optimizer-specific guarantee is inferred here.

\end{document}